\documentclass[11pt]{article}

\usepackage[final]{acl}

\usepackage{float}
\usepackage{times}
\usepackage{latexsym}
\usepackage[T1]{fontenc}
\usepackage[utf8]{inputenc}
\usepackage{microtype}
\usepackage{inconsolata}
\usepackage{graphicx}
\usepackage{booktabs}
\usepackage{amsmath}
\usepackage{multirow}
\usepackage{xcolor}
\usepackage{url}
\usepackage{enumitem}
\usepackage{placeins}
\newcommand{\dec}[1]{{\color{blue}\fontsize{7pt}{7pt}\selectfont\hspace{1pt}\textbf{(-#1)}}}
\newcommand{\inc}[1]{{\color{red}\fontsize{7pt}{7pt}\selectfont\hspace{1pt}\textbf{(+#1)}}}
\newcommand{\same}[1]{{\color{gray}\fontsize{7pt}{7pt}\selectfont\hspace{1pt}(\pm0.000)}}
\usepackage[most]{tcolorbox}
\usepackage{xcolor}
\usepackage{tabularx}
\newtcolorbox{PromptBox}[1]{
  breakable,
  colback=gray!5!white,
  colframe=gray!55!black,
  boxrule=0.5pt,
  arc=2pt,
  left=1.5mm,
  right=1.5mm,
  top=1.2mm,
  bottom=1.2mm,
  title={\small\bfseries #1}
}
\usepackage{array}

\title{Auditing MCQA Benchmarks through Probability Landscapes}

\author{Anonymous ACL submission}

\author{
 \textbf{Minsoo Song\textsuperscript{1}},
 \textbf{Chanjun Park\textsuperscript{1,\textdagger}}
\\
\\
 \textsuperscript{1}Soongsil University 
\\
 \texttt{ecoses042@soongsil.ac.kr, chanjun.park@ssu.ac.kr} 
\\
}
\begin{document}
\maketitle

\begingroup
\renewcommand{\thefootnote}{\textdagger}
\footnotetext{Corresponding authors.}
\endgroup

\begin{abstract}
As Large Language Models rapidly advance, performance on standard
multiple-choice question answering (MCQA) benchmarks is reaching
saturation. While the community has responded by developing increasingly difficult
datasets, validating question quality and filtering flawed items remains a
labor-intensive process. To provide a scalable diagnostic approach, we propose
a two-component probabilistic framework for auditing MCQA benchmarks using
model output distributions. First, for benchmark-level analysis, we characterize the probability landscape
using the top prediction probability ($P_{top1}$) and normalized residual
entropy ($H_{norm}$), summarized globally by Mean Pairwise Distance (MPD).
Second, for item-level diagnostics, we introduce noise injection to reduce
meaningful distractor competition, enabling us to flag candidate items for
targeted human review and categorize residual failure patterns. Across four MCQA benchmarks, our landscape analysis reveals benchmark-level
differences in model confidence and residual option competition. Concurrently,
our noise-injection method flags potentially actionable item-level issues,
showing alignment with expert error annotations from MMLU-Redux. These results
suggest that our probability-based framework provides a lightweight audit lens
for comparing macro-level benchmark structure and prioritizing individual
items for targeted human review.
\end{abstract}

\begin{figure*}[t]
    \centering
    \includegraphics[width=\textwidth]{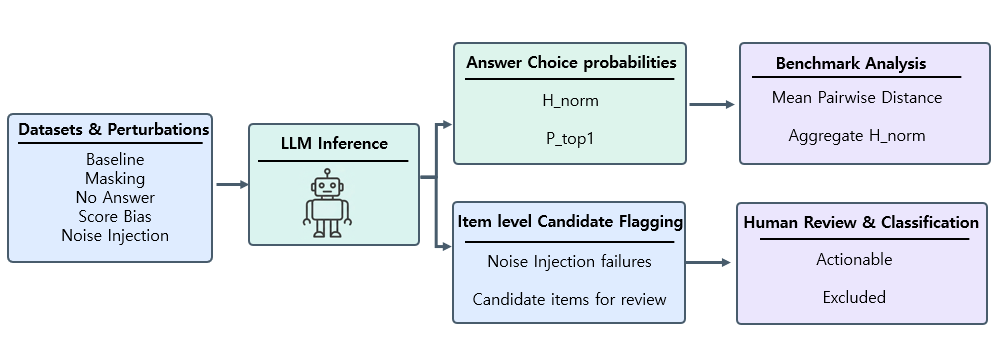}
\caption{Overview of the proposed two-level MCQA auditing framework. After LLM
inference under the baseline and perturbation settings, answer-choice
probabilities support benchmark-level analysis using aggregate $H_{norm}$ and
Mean Pairwise Distance (MPD). In parallel, cross-model noise-injection failures
flag candidate items for targeted human review and coarse classification as
Actionable or Excluded. Candidate flagging prioritizes review rather than
automatically determining benchmark flaws.}
    \label{fig:introduction_diagram}
\end{figure*}
\section{Introduction}

MCQA benchmarks are widely used for evaluating LLMs, in part because their
fixed-choice format enables simple and standardized automatic
scoring~\cite{wei2024rethinking}.

However, as LLM capabilities continue to improve, many widely used benchmarks are approaching performance saturation. Recent studies have shown that frontier models achieve scores close to human level ceilings on several established benchmarks. Prior work has also demonstrated that such improvements may not necessarily reflect genuine reasoning ability, but may instead arise from dataset artifacts or sensitivity to minor prompt variations~\cite{banerjee2024vulnerability, wang2024generalization}.

Recent work has begun to explicitly study the phenomenon of benchmark saturation. For example,~\citet{akhtar2026ai} analyze score trends across multiple frontier models and show that several popular evaluation benchmarks exhibit limited headroom for further performance improvements. Such analyses capture overall performance trends but rely on aggregate accuracy statistics and thus provide limited information about the internal structure of benchmark questions.

To address the saturation problem, recent efforts have focused on constructing more difficult evaluation datasets, such as GPQA~\cite{rein2024gpqa}, Humanity’s Last Exam (HLE)~\cite{phan2025humanity}, and MMLU-Pro~\cite{wang2024mmlu}. These benchmarks aim to resist memorization and shortcut reasoning by increasing problem difficulty and expanding answer spaces. However, building such datasets requires substantial human effort and validation, motivating the need for more cost efficient methods to audit existing benchmarks.

Traditional assessment theory suggests that a well designed multiple choice question should contain plausible distractors that compete meaningfully with the correct answer. In educational measurement, distractor quality is typically evaluated using response distributions collected from human examinees, where strong distractors attract a portion of responses while weak distractors are rarely selected~\cite{kline1999book, haladyna1993many, gierl2017developing}. However, the acquisition of such large-scale response data remains challenging for modern LLM benchmarks.
~\citet{park2023linear} suggests that probabilistic signals derived from model outputs can provide richer insights into model behavior beyond top-1 predictions. In particular, probability distributions over candidate tokens may reflect internal model representations and reveal patterns of uncertainty and competition among alternatives. These signals serve as a proxy for distractor competition using LLM responses.

In this work, we propose a probabilistic framework for auditing MCQA
benchmarks using model output distributions. Instead of relying solely
on accuracy, our approach analyzes how probability mass is distributed
across answer choices to characterize benchmark-level structural
properties and flag items for targeted human review. Figure~\ref{fig:introduction_diagram} illustrates overall outline of our framework.

We introduce Normalized Residual Entropy ($H_{norm}$) to capture the
degree of residual option competition within each question, and analyze
benchmark-level landscape structure using Mean Pairwise Distance (MPD).
We show that aggregate $H_{norm}$ and MPD form descriptive landscape summaries
that reveal broadly consistent benchmark-level patterns across the
well-calibrated models evaluated.

At the item level, our framework follows a cost-efficient two-stage auditing
process. First, noise injection replaces distractors with unrelated city names
to reduce meaningful distractor competition and narrow the benchmark to a
small set of candidate items. Second, only the flagged candidates undergo
targeted human review and are categorized using a taxonomy that separates
perturbation-induced failures from potentially actionable benchmark issues.

We do not treat probability-based metrics or perturbation failures as
definitive evidence of benchmark flaws. Instead, aggregate $H_{norm}$ and MPD
characterize benchmark-level option-competition structure, while
noise-injection failures prioritize candidate items for targeted human review.

Our main contributions are as follows:

\begin{itemize}
    \item We propose a probability-landscape framework for analyzing MCQA
    benchmarks beyond accuracy, representing each item with $P_{top1}$ and
    $H_{norm}$ and summarizing benchmark-level dispersion with MPD.

    \item We show that aggregate $H_{norm}$ and MPD capture complementary
    benchmark-level structures: $H_{norm}$ summarizes model-perceived residual
    option competition, while MPD summarizes the heterogeneity of the
    probability landscape.

    \item We use cross-model noise-injection failures to prioritize items for
    targeted human review, develop a taxonomy that separates perturbation-induced
    failures from potentially actionable benchmark issues, and externally
    evaluate the resulting candidate signals against MMLU-Redux expert
    annotations~\cite{gema2025we}.
\end{itemize}
\section{Related Work}

\subsection{Benchmark Saturation}
Recent work has analyzed benchmark saturation by examining performance differences among top models.~\citet{akhtar2026ai} propose a statistical framework that evaluates whether leading models can still be meaningfully distinguished on existing benchmarks, revealing that many widely used datasets have already lost their discriminative power. However, such approaches primarily rely on leaderboard comparisons and do not directly diagnose the structural causes of saturation within the dataset itself.

\subsection{Dataset Artifacts and Benchmark Validation}
Another line of research investigates the structural validity of benchmark datasets. Prior studies have shown that many MCQA benchmarks contain annotation artifacts that allow models to achieve high accuracy through superficial patterns rather than genuine reasoning~\cite{banerjee2024vulnerability, wang2024generalization}.

Prior work on natural language inference established that annotation artifacts
can enable prediction from incomplete inputs. \citet{gururangan2018annotation}
identified systematic lexical cues introduced during dataset construction,
while \citet{poliak2018hypothesis} demonstrated that hypothesis-only baselines
can exploit such cues to achieve non-trivial performance. Related partial-input
effects have also been identified in MCQA datasets. For example, models can
sometimes predict correct answers even when the question stem is removed,
revealing weak dependencies between questions and answer
choices~\cite{balepur2024artifacts}. Other studies show that the position of the
correct answer can influence model predictions~\cite{zheng2023large}. These
findings motivate diagnostics that test whether benchmark performance depends
on the complete question rather than exploitable structural cues.

MMLU-Redux audits MMLU through expert reannotation, identifying erroneous and
ambiguous benchmark items~\cite{gema2025we}. In contrast to this expert-led
verification, our perturbation-based and classifier-based signals are designed
to automatically prioritize a smaller candidate set for targeted human
review, rather than replace expert auditing. We therefore use the MMLU-Redux
expert annotations as an external validation reference for our candidate
signals.
\subsection{Distractor Quality Evaluation}

Distractor quality has long been studied in educational measurement and psychometrics. Traditional distractor analysis evaluates incorrect answer choices based on how frequently they are selected by examinees in real testing environments. Distractors that are rarely chosen are typically considered ineffective because they fail to represent plausible misconceptions~\cite{kline1999book}. Standard guidelines suggest that each distractor should attract at least a small fraction of examinees, often around five percent, except in very easy questions where the correct answer rate exceeds ninety percent~\cite{haladyna1993many, gierl2017developing}.

Prior work evaluates distractor quality using controlled human responses or QA
models as proxies for learners~\cite{kalpakchi2021bert,luo2024chain,
chung2020bert,offerijns2020better}. These approaches assess whether distractors
meaningfully compete with the correct answer, but require additional response
collection or task-specific evaluation.

\subsection{Probability based Analysis of Model Behavior}

Prior studies have explored probabilistic signals such as model confidence, entropy, and logit differences to analyze the internal behavior of LLMs. Prior work on confidence calibration investigates whether predicted probabilities accurately reflect the correctness of model outputs, revealing that LLMs are often poorly calibrated despite strong predictive performance~\cite{xiong2023can}. 
Other research has examined uncertainty estimation by analyzing distributional statistics of model outputs, including entropy and token level log probabilities. These approaches have been applied to detect unreliable generations and hallucinations by identifying outputs associated with high predictive uncertainty~\cite{farquhar2024detecting}. 

These studies demonstrate that probability distributions contain rich information about model behavior beyond final predictions. However, most prior work focuses on understanding model uncertainty or reliability. In contrast, our work uses probability distributions as a diagnostic signal for auditing benchmark design and characterizing structural properties of MCQA benchmarks.

\section{Methodology} \label{sec:methodology}

We analyze MCQA benchmark structure using probability distributions over answer choices and their changes under controlled perturbations. Our approach consists of two components:
\begin{itemize}
    \item \textbf{Probability Landscape Analysis}: modeling the global structure of MCQA benchmarks using probability distributions derived from model outputs.
\item \textbf{Perturbation-based Diagnostics}: probing benchmark sensitivity
through controlled dataset modifications. Among these, noise injection serves
as the primary probe for item flagging, while masking, no-answer
injection, and score bias characterize benchmark-level robustness.
\end{itemize}
Figure~\ref{fig:introduction_diagram} provides an overview of the proposed diagnostic pipeline.
\subsection{Probabilistic Representation}

For each question, the model produces a probability distribution over answer choices.

Let $\mathcal{O}=\{x_1,\dots,x_k\}$ denote the set of options. 
Given a question $q$, the model outputs

\[
P(x_i \mid q)
\]

for each $x_i \in \mathcal{O}$, and predicts

\[
y_{pred} = \arg\max_{x_i \in \mathcal{O}} P(x_i \mid q).
\]

These probabilities allow analysis of competition among answer choices beyond the final prediction.

\subsection{Distractor Structure Representation}

A well-designed MCQA question should present several plausible distractors that compete with the correct answer. If the model assigns most residual probability mass to a single non-selected option, the item exhibits concentrated residual competition from the model's perspective. This pattern may indicate weak distractor competition, but it is not by itself sufficient evidence of poor item quality.

To represent this property, we characterize each question using two probabilistic features.

\paragraph{Model Confidence}

Model confidence is defined as the probability assigned to the top prediction:
\[
P_{top1}
\]
This value captures how strongly the model favors its predicted answer.

\paragraph{Normalized Residual Entropy}

To measure competition among distractors, we define \textit{Normalized Residual Entropy} ($H_{norm}$) over the non selected options.

\begin{equation}
\resizebox{0.8\hsize}{!}{$
H_{norm} = \frac{-\sum_{i \in \mathcal{O}\setminus\{y_{pred}\}} P'(x_i)\log_2 P'(x_i)}{\log_2(|\mathcal{O}|-1)}
$}
\end{equation}

where $P'$ denotes the renormalized probability distribution over the remaining options.

$H_{norm}$ ranges from $0$ to $1$, where larger values indicate stronger competition among distractors. In addition, questions with extremely low residual entropy ($H_{norm} \approx 0$) indicate highly concentrated residual competition, where one non-selected option dominates the remaining probability mass.
Together, $P_{top1}$ and $H_{norm}$ form a compact representation of each
question's probabilistic structure, enabling comparison of distractor
competition patterns across benchmarks and models.

Each question can therefore be represented as a two dimensional vector

\[
z_i = (P_{top1}^{(i)}, H_{norm}^{(i)})
\]

which captures both answer confidence and distractor competition.

\subsection{Probability Landscape Cohesion}

The set of question representations forms a \textit{probability landscape} describing how model predictions are distributed across a benchmark.

To quantify the global structure of this landscape, we compute the MPD:
\begin{equation}
\resizebox{0.8\hsize}{!}{$
MPD = \frac{2}{N(N-1)} \sum_{i<j} \| z_i - z_j \|_2
$}
\end{equation}

where $N$ is the number of questions and $\|\cdot\|_2$ denotes Euclidean distance.

For the main benchmark-level analysis, we pool the 1,000 item representations
from each of the three small-scale models, yielding $N=3{,}000$
$(\text{item},\text{model})$ points per benchmark. The resulting MPD includes
both within-model and cross-model pairs and therefore differs from the mean of
the independently computed per-model MPDs in Appendix~\ref{app:model_detail}.

Lower MPD values indicate that items occupy a narrower region of the probability landscape, whereas higher MPD values indicate greater heterogeneity in model confidence and residual option competition. We use MPD as a descriptive summary of landscape dispersion, not as a direct measure of benchmark quality.

\subsection{Diagnostic Perturbations}

To probe benchmark structure, we introduce a set of controlled perturbations
that modify the dataset while preserving the overall task format.
Three perturbations: token masking, no-answer injection, and score bias are analyzed in Section~\ref{sec:benchmark_char}.
A fourth perturbation, noise injection, serves as the primary probe for
item flagging and is analyzed in depth in Section~\ref{sec:taxonomy}.

\paragraph{Token Masking}

To examine the dependence on question context, we construct masked variants in which all tokens in the question body are replaced with a special token (e.g., \texttt{[MASK]}).

Replacing rather than deleting the question preserves the prompt template and
question slot while removing semantic content. This isolates dependence on
question information without introducing a structural change to the input
format.

If models maintain high accuracy under this condition, it suggests that answers can be inferred from option patterns rather than from the question content.

\paragraph{No-Answer setting}

To examine the forced choice bias inherent in MCQA evaluation, we replace the original correct answer with a special option labeled ``No Answer''.

This setting evaluates whether models can recognize the absence of a valid solution rather than selecting the most plausible distractor.

\paragraph{Score Bias}

To evaluate sensitivity to irrelevant textual signals, we append the phrase ``(Score: 10/10)'' to the end of each question.

We use \texttt{(Score: 10/10)} as the representative irrelevant evaluative
suffix; Appendix~\ref{app:score_bias_values} shows that accuracy remains stable
across alternative score values within the same template.

If benchmark structure is robust, this modification should have minimal impact on model predictions or probability landscapes.
Significant shifts in MPD or prediction accuracy indicate sensitivity to superficial textual cues rather than the semantic content of the question.

\paragraph{Noise Injection}

To eliminate semantic competition between distractors, we replace the original distractors with semantically unrelated tokens. 
Specifically, distractors are replaced with city names selected heuristically to ensure semantic independence from the question context.

Under this condition, the correct answer should become trivially identifiable.
Questions that remain incorrect even after noise injection are treated as
flagged items for further inspection. Such failures may arise from ambiguity,
answer-key issues, or preprocessing artifacts, but they may also reflect
limitations of the perturbation itself.
Failures under noise injection are not uniformly distributed but cluster into
identifiable types reflecting systematic properties of benchmark items.
\section{Experimental Setup}

\subsection{Models}

We evaluate three open-weight instruction-tuned models below 10B parameters:
Qwen-2.5 7B~\cite{qwen25technicalreport}, Gemma-2 9B~\cite{gemma2024gemma}, and
Llama-3 8B~\cite{llama3modelcard}. Supplementary scalability experiments use
Qwen-2.5 32B, Qwen-2.5 72B, and Llama-3.1 70B. Model and inference details are
reported in Appendix~\ref{app:model_inference_details}, with scaling results in
Appendix~\ref{app:scaling_effects}.
\begin{figure*}[t]
    \centering
    \includegraphics[width=\textwidth]{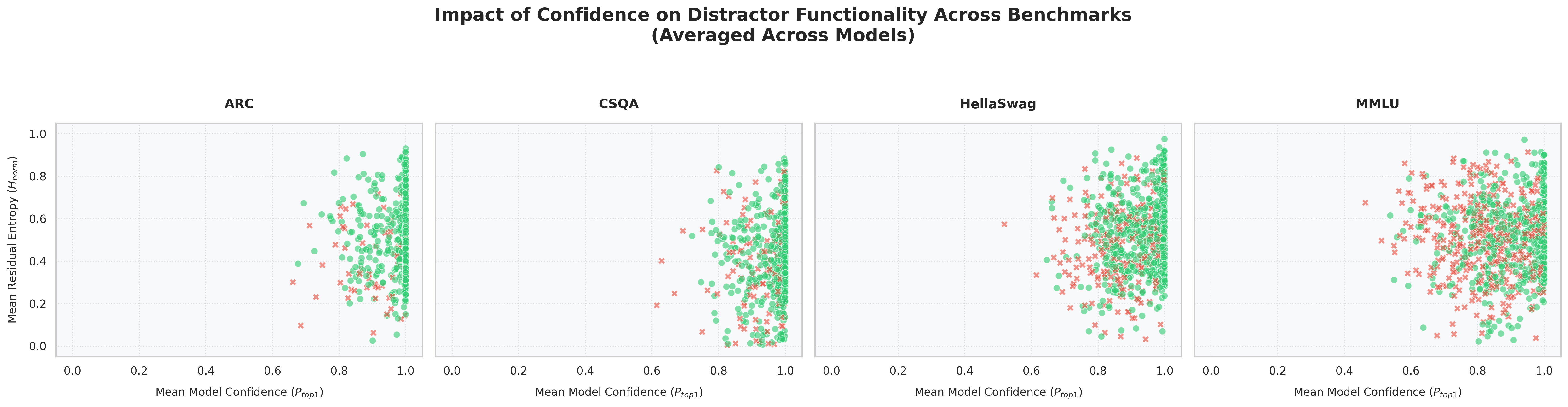}
    \caption{
    Distribution of $H_{norm}$ versus $P_{top1}$ for 1,000 questions per benchmark, with coordinates averaged across three models.
    ARC and CSQA concentrate near $P_{top1} \approx 1.0$, reflecting high model confidence on most items;
    MMLU and HellaSwag span a wider region, consistent with greater diversity in item difficulty.
    }
    \label{fig:entropy_scatter}
\end{figure*}

\subsection{Benchmarks}

We evaluate our framework on four widely used MCQA benchmarks: MMLU~\cite{hendrycks2020measuring}, ARC~\cite{clark2018think}, HellaSwag~\cite{zellers-etal-2019-hellaswag}, and CommonsenseQA (CSQA)~\cite{talmor-etal-2019-commonsenseqa}.

For each dataset, we randomly sample 1,000 instances from the available splits to obtain a computationally manageable yet representative evaluation set.

To ensure consistency across benchmarks, we standardize the five-option CSQA
items to a four-option format using a deterministic positional rule. We remove
option E by default; when E is the correct answer, we instead remove option D
and relabel the original option E as D. This procedure avoids removing the
correct answer and does not depend on random or plausibility-based distractor
selection. We additionally evaluate the original five-choice format in
Appendix~\ref{app:5choice} to assess the effect of choice-set size on
probability-landscape structure.

\subsection{Prompting and Probability Extraction}
Each question is formatted using a standardized multiple choice template with four options (A–D). A system instruction forces the model to output only the answer letter (A–D), enabling direct extraction of logits for each option token. The prompt template and invalid output statistics are reported in Appendix section~\ref{app:prompt_template}.

All experiments use deterministic decoding (temperature = 0, top-p = 1) with the maximum generation length limited to one token. No logit bias or additional constraints are applied.

Instead of evaluating generated text, we analyze the probability distribution over answer choices. Logits for the tokens \texttt{A}, \texttt{B}, \texttt{C}, and \texttt{D} at the final prompt position are extracted and converted into probabilities using the Softmax function.

\section{Benchmark-Level Probability Landscape Analysis} \label{sec:benchmark_char}
\begin{table}[t]
\centering
\begin{tabular}{lcc}
\toprule
Benchmark & Mean $H_{norm}$ & MPD \\
\midrule
ARC       & 0.572 & 0.404 \\
HellaSwag & 0.552 & 0.433 \\
MMLU      & 0.536 & 0.479 \\
CSQA      & 0.488 & 0.424 \\
\bottomrule
\end{tabular}
\caption{Benchmark-level mean $H_{norm}$ and pooled MPD across three
small-scale models ($N=3{,}000$ item--model points per benchmark).}
\label{tab:hnorm_aggregate}
\end{table}
\begin{table*}[t]
\centering
\begin{tabular}{lccccc}
\hline
Benchmark & Baseline & A bias & B bias & C bias & D bias \\
\hline
ARC       & 0.404 & 0.357 \dec{0.047} & 0.419 \inc{0.015} & 0.424 \inc{0.020} & 0.357 \dec{0.047} \\
CSQA      & 0.424 & 0.410 \dec{0.014} & 0.421 \dec{0.003} & 0.431 \inc{0.007} & 0.415 \dec{0.009} \\
HellaSwag & 0.433 & 0.427 \dec{0.006} & 0.432 \dec{0.001} & 0.435 \inc{0.002} & 0.423 \dec{0.010} \\
MMLU      & \textbf{0.479} & \textbf{0.466} \dec{0.013} & \textbf{0.470} \dec{0.009} & \textbf{0.486} \inc{0.007} & \textbf{0.470} \dec{0.009} \\
\hline
\end{tabular}
\caption{
MPD of probability distributions under answer-position bias conditions.
Each bias condition forces the correct answer to appear at a fixed option position (A–D).
Values in parentheses indicate the change relative to the baseline setting.
Overall, MPD remains relatively stable across positional bias conditions, suggesting that the global probability landscape is largely robust to answer-position perturbations.
}
\label{tab:position_bias_mpd}
\end{table*}
\begin{table*}[t]
    \centering
    \setlength{\tabcolsep}{6pt}
    \begin{tabular}{lrrrrrr}
        \toprule
        Benchmark & Baseline & Mask(100\%) & No Answer & Noise Injection & Score Bias(10)\\
        \midrule
        ARC & {91.4}  & 40.5 \dec{50.9} & {40.8} \dec{50.6} & {99.7} \inc{8.3}  &  {91.4}\\
        CSQA & 83.4 & 26.7 \dec{56.8} & 32.4 \dec{51.0} & 99.1 \inc{15.6}  & 83.2 \dec{0.2}\\
        HellaSwag & 73.3 & {58.0} \dec{15.3} & 36.5 \dec{36.8} & 98.5 \inc{25.2} & 73.6 \inc{0.3} \\
        MMLU & 60.2  & 32.5 \dec{27.7} & 17.1 \dec{43.1} & 98.6 \inc{37.7}  & 59.8 \dec{0.4}\\
        \bottomrule
    \end{tabular}
    \caption{
    Accuracy (\%) across perturbation conditions.
    Masking and no-answer settings significantly reduce accuracy, while noise injection greatly increases performance by removing semantic competition among distractors.
    Score bias produces minimal changes relative to the baseline.
    }
    \label{tab:accuracy_results}
\end{table*}
\subsection{Baseline Probability Landscape}

We analyze baseline probability landscapes of MCQA benchmarks using the joint distribution of $P_{top1}$ and $H_{norm}$, where each question is represented as a point in this space. 

Figure~\ref{fig:entropy_scatter} shows that ARC and CSQA exhibit concentrated distributions in the high-confidence region ($P_{top1} \approx 1.0$), indicating that models often assign near-deterministic probabilities to predicted answers. 

To quantify landscape structure, we measure cohesion using Mean Pairwise Distance (MPD). Lower MPD values indicate more concentrated landscapes, whereas higher values reflect greater structural diversity across questions. 

Table~\ref{tab:hnorm_aggregate} shows clear differences across benchmarks: ARC exhibits the most concentrated landscape (MPD = 0.404), while MMLU shows a more dispersed structure (MPD = 0.479). This ordering---ARC $<$ CSQA $<$ HellaSwag $<$ MMLU in MPD---is also examined at the model level in Appendix~\ref{app:model_detail}.
As a supplementary extension to a more challenging benchmark, we apply the
same pooled MPD aggregation to existing baseline outputs for all 448 GPQA items
across three models. The resulting pooled MPD is 0.477 ($N=1{,}344$), as
reported in Appendix~\ref{app:gpqa_baseline}.

\subsection{Aggregate \texorpdfstring{$H_{norm}$}{H\_norm} and MPD Patterns}

Mean $H_{norm}$ and MPD provide complementary aggregate summaries of option
competition and landscape heterogeneity, respectively. Their rankings can
differ: MMLU has the highest MPD despite moderate mean $H_{norm}$. Because
item-level $H_{norm}$ correlations across models are near zero
(Appendix~\ref{app:hnorm_consistency}), we restrict these metrics to
benchmark-level characterization and use cross-model noise-injection failures
for item-level candidate prioritization.

\subsection{Robustness to Positional Bias}

Table~\ref{tab:position_bias_mpd} shows that MPD is generally stable across
answer-position perturbations. ARC is the main exception, decreasing from
0.404 to 0.357 under the A- and D-biased settings, while the other benchmarks
show smaller shifts.

\subsection{Perturbation-Based Audit Signals}

To further investigate benchmark structure, we apply several controlled perturbations that modify either the question context or the answer set. Table~\ref{tab:accuracy_results} reports the accuracy changes across perturbation settings, while Table~\ref{tab:MPD_perturbation} shows the corresponding MPD changes.

\begin{table*}[t]
\centering
\begin{tabular}{lrrrr}
\hline
Benchmark & Baseline & Mask(100\%) & No Answer & Score Bias(10) \\ \hline
ARC       & 0.404 & 0.427 \inc{0.023} & 0.424 \inc{0.020} & 0.357 \dec{0.047} \\
CSQA      & 0.424 & 0.395 \dec{0.029} & 0.431 \inc{0.007} & 0.403 \dec{0.021} \\
HellaSwag & 0.433 & 0.423 \dec{0.010} & 0.445 \inc{0.012} & 0.425 \dec{0.008} \\
MMLU      & 0.479 & 0.419 \dec{0.060} & 0.452 \dec{0.027} & 0.462 \dec{0.017} \\ \hline
\end{tabular}
\caption{
MPD under different perturbation settings, computed over the pooled
$(\text{item},\text{model})$ points from the three small-scale models.
Values in parentheses represent the change relative to the baseline.
Despite dataset perturbations such as masking and no-answer insertion, MPD varies only moderately across benchmarks, indicating that the global probability landscape is relatively robust to these modifications.
}
\label{tab:MPD_perturbation}
\end{table*}
\begin{table*}[!t]
  \centering
  \scriptsize
  \setlength{\tabcolsep}{3.5pt}
  \renewcommand{\arraystretch}{0.88}
  \begin{tabular}{@{}p{0.12\linewidth}p{0.19\linewidth}p{0.57\linewidth}r@{}}
  \toprule
  \textbf{Group} & \textbf{Type} & \textbf{Description} & \textbf{Count} \\
  \midrule
  \multirow{2}{*}{Excluded}
  & X1. Negation / EXCEPT
  & Injected city names also satisfy the negated condition, destroying the original contrastive option set. & 15 \\
  & X2. Semantic collision
  & Injected city names accidentally become plausible answers, especially in location-related questions. & 6 \\
  \midrule
  \multirow{5}{*}{Actionable}
  & A1. Scope inconsistency
  & The question condition conflicts with the labeled answer, e.g., asking about ``all'' cases while the answer applies only to a subset. & 1 \\
  & A2. Answer-key error
  & The labeled answer is factually incorrect, or another option is more correct. & 3 \\
  & A3. Encoding artifact
  & Option text is corrupted by preprocessing or file conversion, such as fractions converted into date strings. & 1 \\
  & A4. Narrative ambiguity
  & The context does not uniquely determine the intended continuation, leaving multiple options plausible. & 3 \\
  & A5. Trivia-dependent underspecification
  & The answer depends on culture- or episode-specific knowledge not inferable from the question alone. & 1 \\
  \bottomrule
  \end{tabular}
  \caption{
  Taxonomy of residual failures under noise injection. Excluded types reflect
  perturbation limitations, while actionable types identify candidate items for
  targeted human review. Counts use the final four-benchmark scope.
  }
  \label{tab:noise_taxonomy}
\end{table*}

Masking and no-answer insertion substantially reduce accuracy, whereas noise
injection raises accuracy to nearly 100\%, confirming that most errors disappear
when meaningful distractor competition is removed. Relative no-answer accuracy
drops are largest for MMLU (71.6\%), followed by CSQA (61.2\%), ARC (55.4\%),
and HellaSwag (50.2\%).

Score bias produces only minor accuracy changes but alters MPD in some
benchmarks. We use \texttt{(Score: 10/10)} as the representative value, with
the score-value comparison reported in
Appendix~\ref{app:score_bias_values}.

\section{Noise-Injection Failure Analysis}
\label{sec:taxonomy}
This section focuses on categorizing these residual failures derived from the noise injection procedure.

\subsection{Taxonomy of Residual Failures}

We analyze residual failures using a two-stage procedure: candidates are first
identified through cross-model noise-injection failures and then undergo
targeted human review. Table~\ref{tab:noise_taxonomy} groups the reviewed items
into Excluded perturbation-induced cases and potentially Actionable benchmark
issues; fine-grained subtypes are used only for qualitative analysis.
Inter-annotator agreement is reported in
Appendix~\ref{app:taxonomy_agreement}.

Of the 30 residual failures, 21 are Excluded and 9 are retained as Actionable
candidates. The X2 count is 6 in the final four-benchmark scope; the earlier
11/36 count included the separate five-choice CSQA analysis. Representative
examples are provided in Appendix~\ref{app:taxonomy_examples}.

\subsection{Complementary Structural Classification}

As a complementary signal, we apply an MMLU-Redux-style GPT-4o-mini classifier
that labels items as \texttt{ok}, \texttt{bad\_question\_clarity}, or
\texttt{bad\_options\_clarity}. It flags 18.4\% of CommonsenseQA and 2.8\% of
MMLU, providing additional candidates for targeted human review. Full
classification results and implementation details are reported in
Appendix~\ref{app:llm_classification}.
\begin{table}[t]
\centering
\small
\resizebox{\columnwidth}{!}{
\begin{tabular}{lrrrrrr}
\toprule
Signal & Flagged & Recall & Precision & F1 & Enrichment & $p$ \\
\midrule
Noise injection
& 2 & 0.222 & 1.000 & 0.364 & 41.4$\times$ & 0.0005 \\
Noise injection $\cup$ LLM classifier
& 11 & 0.444 & 0.364 & 0.400 & 15.1$\times$ & $<0.0001$ \\
Random baseline
& -- & -- & 0.024 & -- & 1.0$\times$ & -- \\
\bottomrule
\end{tabular}
}
\caption{Validation of the candidate signals against expert-identified flaws
in the 373-item MMLU-Redux overlap.}
\label{tab:mmlu_redux_validation}
\end{table}
\subsection{Validation against MMLU-Redux Expert Labels}

We match our 1,000-item MMLU subset against MMLU-Redux by exact question text,
yielding 373 overlapping items, of which 9 are labeled as flawed by expert
review. 

Table~\ref{tab:mmlu_redux_validation} shows that the proposed signals
substantially enrich expert-identified flaws relative to
random inspection, but recall remains low (0.222--0.444). We therefore
interpret them as candidate-prioritization signals for targeted human review rather than
as exhaustive flaw detectors. Because only 9 confirmed flaws occur in the
overlap, the exact precision and recall estimates should be interpreted as
directional.

\section{Conclusion}
As LLM performance on MCQA benchmarks approaches saturation, accuracy alone becomes insufficient for assessing benchmark quality. This motivates systematic methods for auditing the structural properties of existing evaluation datasets.

In this work, we proposed a probabilistic framework for auditing MCQA benchmarks using model output distributions. At the benchmark level, aggregate $H_{norm}$ and MPD provide descriptive summaries of probability-landscape structure, with broadly consistent benchmark-level patterns across the well-calibrated models evaluated. At the item level, noise injection serves as a first-pass candidate-prioritization signal: it narrows the set of items requiring targeted human review, after which residual failures can be separated into perturbation-induced cases and potentially actionable benchmark issues. Validation against MMLU-Redux expert annotations further shows that these signals enrich flawed items relative to random inspection, although their low recall precludes their use as exhaustive flaw detectors.

Overall, our framework offers a practical foundation for benchmark auditing by providing two complementary signals: aggregate landscape characterization for comparing benchmark design, and item-level flagging for prioritizing questions for targeted human review.

\section*{Limitations}

This work has several limitations. First, our experiments do not include the
latest frontier or closed-source LLMs, and the larger models are used only as
supplementary scalability checks. Thus, the findings should be interpreted as
evidence from an open-model setting rather than as fully general conclusions
about all current LLMs.

Second, the proposed metrics rely on model output probabilities and may be
affected by calibration, tokenization, prompting, quantization, inference
backend, and answer-letter probability extraction. Accordingly, $H_{norm}$ and
MPD are descriptive model-based audit signals, not direct measures of benchmark
quality.

Third, $H_{norm}$ is limited as an item-level diagnostic. Since item-level
$H_{norm}$ rankings are unstable across models, low $H_{norm}$ should not be
treated as automatic evidence of flawed questions. We therefore use $H_{norm}$
mainly for aggregate benchmark characterization.

Fourth, our external item-level validation is limited to 373 exact-text matches
with MMLU-Redux, containing only 9 expert-identified flaws. Although the
proposed signals substantially enrich flawed items relative to random
inspection, their recall remains low (0.222--0.444), and estimates based on
only 9 positive cases are necessarily unstable. The signals should therefore
be interpreted as first-pass candidate-prioritization signals for targeted human review
rather than exhaustive or automatic benchmark-flaw detectors. In addition,
the double-annotated taxonomy subsets are small, and the weaker fine-grained
agreement observed for HellaSwag limits conclusions about the reliability of
individual subtypes.

Finally, our primary empirical scope is limited to four MCQA benchmarks, with
GPQA included only as a supplementary baseline extension. The primary
experiments use sampled subsets of 1,000 instances per benchmark and
preprocessing choices such as the four-choice standardization of CommonsenseQA.
Future work should evaluate full benchmark splits, broader model families,
frontier models, human annotations, and uncertainty estimates for aggregate
statistics such as MPD.

\section*{Ethical Statement}

This work analyzes publicly available MCQA benchmarks and model outputs. No new datasets involving human participants were collected, and no personal or sensitive information was used. Our study focuses on diagnosing structural properties of benchmarks rather than deploying models in real world applications. We believe that improving benchmark design can contribute to more reliable and transparent evaluation of language models.
\section*{Acknowledgments}
This work was supported by the Korea Internet \& Security Agency (KISA) grant funded by the Korea government (PIPC) (No. RS-2026-25526342, Development of Technologies for Preventing Sensitive Information Inference and Risk Assessment in Foundation Model Operations). This work was also supported by the National Research Foundation of Korea (NRF) grant funded by the Korea government (MSIT) (RS-2026-25483747). This research was further supported by the Culture, Sports and Tourism R\&D Program through the Korea Creative Content Agency, funded by the Ministry of Culture, Sports and Tourism in 2026 (Project Name: Develop AI Agent Technology to Connect Knowledge through Public Cultural Facility-Based Discussion and Communication, Project Number: RS-2026-25520645).
    
\bibliography{custom}

\clearpage
\newpage
\appendix

\section{Model-level MPD Detail}
\label{app:model_detail}
\begin{table}[h]
\centering
\small
\setlength{\tabcolsep}{3.5pt}
\resizebox{\columnwidth}{!}{
\begin{tabular}{lcccc}
\toprule
Model & ARC & CSQA & HellaSwag & MMLU \\
\midrule
Gemma-2 9B        & 0.169 & 0.285 & 0.274 & \textbf{0.369} \\
Llama-3 8B        & 0.376 & 0.384 & 0.414 & \textbf{0.450} \\
Qwen-2.5 7B       & 0.374 & 0.360 & 0.381 & \textbf{0.385} \\
\midrule
Llama-3.1 70B     & 0.298 & 0.363 & \textbf{0.367} & 0.336 \\
Qwen-2.5 32B      & 0.071 & 0.123 & 0.204 & \textbf{0.187} \\
Qwen-2.5 72B      & 0.335 & 0.278 & 0.354 & \textbf{0.398} \\
\bottomrule
\end{tabular}
}
\caption{Per-model MPD values by benchmark. Upper block: small-scale models used
in main experiments. Lower block: large-scale models evaluated for scalability.}
\label{tab:mpd_by_model}
\end{table}
Table~\ref{tab:mpd_by_model} reports per-model MPD values across all four benchmarks
for all six evaluated models.
The three small-scale models (Gemma-2 9B, Llama-3 8B, Qwen-2.5 7B)
consistently place MMLU at the top of the MPD ranking,
showing that this ordering is not unique to a single small-scale model. This
agreement does not establish that the ordering is independent of model-specific
calibration or inference characteristics.
\begin{table}[h]
\centering
\small
\setlength{\tabcolsep}{3.5pt}
\resizebox{\columnwidth}{!}{
\begin{tabular}{lcccc}
\toprule
Model & 1st & 2nd & 3rd & 4th \\
\midrule
Gemma-2 9B        & MMLU & CSQA      & HellaSwag & ARC  \\
Llama-3 8B        & MMLU & HellaSwag & CSQA      & ARC  \\
Qwen-2.5 7B       & MMLU & HellaSwag & ARC       & CSQA \\
Qwen-2.5 72B      & MMLU & HellaSwag & ARC       & CSQA \\
\midrule
Llama-3.1 70B     & HellaSwag & CSQA  & MMLU      & ARC  \\
Qwen-2.5 32B      & HellaSwag & MMLU  & CSQA      & ARC  \\
\bottomrule
\end{tabular}
}
\caption{Benchmark MPD rank order per model.}
\label{tab:mpd_rank}
\end{table}

Table~\ref{tab:mpd_rank} summarizes the benchmark MPD rank order per model.
Among the three small-scale models (Gemma-2 9B, Llama-3 8B, Qwen-2.5 7B),
MMLU consistently ranks first.
This consistency across architecturally distinct models suggests that the
ordering is not driven by any single model, although the evidence remains
limited to the evaluated models.

\section{Selection of the Score-Bias Prompt}
\label{app:score_bias_values}

The score-bias perturbation appends an irrelevant evaluative suffix to each
question. To verify that its behavior was not tied to an arbitrary displayed
score, we used the same parenthetical template while varying the value among
\texttt{1/10}, \texttt{10/10}, and \texttt{100/100}. These are value variants
of one fixed textual format rather than distinct formatting templates.

Across the 20 model--benchmark combinations for which all three values are
available, the maximum accuracy spread is 1.10 percentage points, and most
spreads are within one point. Because the differences across these score-value
settings are small, we select \texttt{(Score: 10/10)} as the representative
score-bias prompt in the main experiments.

\begin{table}[h]
\centering
\small
\begin{tabular}{lr}
\toprule
Selection evidence & Reported result \\
\midrule
Values tested & 1/10, 10/10, 100/100 \\
Complete three-value comparisons & 20 \\
Maximum accuracy spread & 1.10 pp \\
Typical accuracy spread & $\leq 1.00$ pp \\
Representative value & 10/10 \\
\bottomrule
\end{tabular}
\caption{Evidence used to select \texttt{(Score: 10/10)} as the representative
score-bias prompt. All tested values use the same parenthetical template, and
the comparison is based on accuracy.}
\label{tab:score_bias_value_summary}
\end{table}

The largest observed spread occurs for Qwen-2.5 32B on MMLU. The purpose of
this comparison is only to justify the representative prompt value; the full
score-bias analysis in the main text uses \texttt{(Score: 10/10)}.

\section{Model and Inference Details} \label{app:model_inference_details}
\begin{table}[h]
\centering
\small
\setlength{\tabcolsep}{3pt}
\begin{tabular}{@{}lcccc@{}}
\toprule
\textbf{Model} & \textbf{Params} & \textbf{Ctx.} & \textbf{Emb.} & \textbf{Quant.} \\
\midrule
Qwen-2.5 7B        & 7.6B  & 32,768  & 3,584 & Q4\_K\_M \\
Gemma-2 9B         & 9.2B  & 8,192   & 3,584 & Q4\_0 \\
Llama-3 8B         & 8.0B  & 8,192   & 4,096 & Q4\_0 \\
Qwen-2.5 32B       & 32.8B & 32,768  & 5,120 & Q4\_K\_M \\
Qwen-2.5 72B       & 72.7B & 32,768  & 8,192 & Q4\_K\_M \\
Llama-3.1 70B      & 70.6B & 131,072 & 8,192 & Q4\_K\_M \\
\bottomrule
\end{tabular}
\caption{Architectural specifications of the evaluated models.}
\label{tab:model_specs}
\end{table}
The four primary benchmark experiments are conducted using the Ollama inference framework with the \texttt{llama.cpp} backend. Table~\ref{tab:model_specs} summarizes the models used in our experiments.
For all models, we use deterministic decoding settings to eliminate stochastic variation in outputs. Specifically, we set temperature to 0, top-p to 1, and limit the maximum number of generated tokens to 1. No logit bias or token level constraints are applied.
 The small models were executed locally using Ollama (v0.17.7), which uses llama.cpp as the inference backend. 
The larger models were evaluated on Google Colab with A100 GPUs using the same runtime and backend configuration.

\section{Scaling Effects}\label{app:scaling_effects}
\begin{table}[H]
\centering
\small
\begin{tabular}{lcc}
\hline
Benchmark & Llama-3 8B & Llama-3.1 70B \\ 
\hline
ARC       & \textbf{0.376} & 0.298  \\
CSQA      & \textbf{0.384} & 0.363  \\
HellaSwag & \textbf{0.414}& 0.367  \\
MMLU      & \textbf{0.450} & 0.410  \\
\hline
\end{tabular}
\caption{
MPD across Llama model scales.
The 70B model consistently produces lower MPD than the 8B model across all benchmarks, suggesting that larger models assign sharper, more concentrated probability distributions over answer choices.
}
\label{tab:llama_scale}
\end{table}

\begin{table}[H]
\setlength{\tabcolsep}{3pt}
\centering
\small
\begin{tabular}{lccc}
\hline
Benchmark & Qwen-2.5 7B & Qwen-2.5 32B & Qwen-2.5 72B \\ 
\hline
ARC       & \textbf{0.374} & 0.071  & 0.335  \\
CSQA      & \textbf{0.360}& 0.123  & 0.278  \\
HellaSwag & \textbf{0.381} & 0.204  & 0.354  \\
MMLU      & \textbf{0.385} & 0.212  & 0.364  \\
\hline
\end{tabular}
\caption{
MPD across Qwen-2.5 model scales.
While the 72B model generally produces lower MPD than the 7B model, the 32B model exhibits substantially lower MPD than both, deviating from a monotonic scaling trend and suggesting model-specific factors beyond raw parameter count.
}
\label{tab:qwen_scale}
\end{table}

Across model families, larger models generally produce lower MPD values, 
indicating a more concentrated probability landscape. 
Table~\ref{tab:llama_scale} shows that the 70B model 
consistently produces lower MPD values than the 8B model across all benchmarks.

However, the scaling behavior is not strictly monotonic. 
In Table~\ref{tab:qwen_scale}, the 32B model exhibits substantially lower MPD values 
than both the 7B and 72B variants. 
This deviation from monotonic scaling suggests that the structure of the 
probability landscape may depend not only on model capacity but also on 
model specific architectural or training characteristics.

Despite these variations, the benchmark-level ordering remains broadly similar
for the well-calibrated models evaluated, while deviations for individual
models indicate sensitivity to model-specific calibration and inference
characteristics.


\section{CommonsenseQA: 4-Choice vs.\ 5-Choice} \label{app:5choice}

We compare the probability landscape under the original 5-choice CommonsenseQA format against the 4-choice version used in the main experiments, using the three small-scale models (Gemma-2 9B, Llama-3 8B, Qwen-2.5 7B).
Table~\ref{tab:5choice_comparison} reports accuracy, $H_{norm}$, and $P_{top1}$ under the 4-choice and original 5-choice CommonsenseQA formats.

\begin{table}[H]
\centering
\small
\begin{tabular}{llrrr}
\toprule
Model & $n$ & Acc (\%) & $H_{norm}$ & $P_{top1}$ \\
\midrule
Gemma-2 9B   & 4-ch & \textbf{85.3} & \textbf{0.754} & \textbf{0.980} \\
             & 5-ch & 81.1 & 0.725 & 0.974 \\
\midrule
Llama-3 8B   & 4-ch & \textbf{78.5} & 0.390 & \textbf{0.945} \\
             & 5-ch & 76.9 & \textbf{0.604} & 0.873 \\
\midrule
Qwen-2.5 7B  & 4-ch & \textbf{86.5} & 0.320 & \textbf{0.988} \\
             & 5-ch & 83.4 & 0.320 & 0.985 \\
\bottomrule
\end{tabular}
\caption{Accuracy, $H_{norm}$, and $P_{top1}$ for CommonsenseQA under 4-choice and 5-choice formats (1000 items, baseline condition).}
\label{tab:5choice_comparison}
\end{table}

Accuracy decreases uniformly across all three models when a fifth choice is introduced, consistent with the additional option making the task more difficult. The effect on $H_{norm}$, however, diverges by model. Gemma-2 9B shows a slight decrease ($-$0.029), consistent with the fifth option receiving negligible residual probability. Llama-3 8B exhibits a large increase ($+$0.213), indicating that the added option substantially redistributes residual probability for most items and reflects calibration instability under the 5-choice format. Qwen-2.5 7B shows no change ($+$0.000), suggesting the most stable probability allocation across choice set sizes.
The benchmark-level $H_{norm}$ rank ordering is preserved for the two well-calibrated models: both Gemma-2 9B and Qwen-2.5 7B maintain the same CSQA rank relative to the other benchmarks. Llama-3 8B is the exception, where the $H_{norm}$ surge in CSQA disrupts the rank ordering. This pattern suggests that the aggregate ordering can be reproducible among well-calibrated models while remaining sensitive to model-specific calibration.

\section{Prompt Template} \label{app:prompt_template}

Each question is formatted using a standardized multiple choice template consisting of the question followed by four answer choices (A–D). The prompt used for evaluation is structured as follows:

\begin{verbatim}
Question: {question}

A. {choice_A}
B. {choice_B}
C. {choice_C}
D. {choice_D}

Answer:
\end{verbatim}

To ensure consistent response formatting across models, we apply the following system instruction:

\begin{quote}
\texttt{You are a helpful assistant. Output only the answer choice letter (A, B, C, or D) and nothing else.}
\end{quote}

This instruction encourages the model to produce a single token corresponding to one of the answer choices, which enables direct extraction of logits associated with each choice. Table~\ref{tab:response_format_validity} shows statistics for invalid form of model responses.
\begin{table*}[h]
\centering
\small
\begin{tabular}{llrrrrr}
\toprule
Model & Dataset & Total & Valid & Format Err & Invalid & Valid Ratio \\
\midrule
Llama-3 8B & ARC & 1000 & 1000 & 0 & 0 & 1.000 \\
Llama-3 8B & CSQA & 1000 & 999 & 0 & 1 & 0.999 \\
Llama-3 8B & HellaSwag & 1000 & 1000 & 0 & 0 & 1.000 \\
Llama-3 8B & MMLU & 1000 & 1000 & 0 & 0 & 1.000 \\

Gemma-2 9B & ARC & 1000 & 1000 & 0 & 0 & 1.000 \\
Gemma-2 9B & CSQA & 1000 & 1000 & 0 & 0 & 1.000 \\
Gemma-2 9B & HellaSwag & 1000 & 1000 & 0 & 0 & 1.000 \\
Gemma-2 9B & MMLU & 1000 & 1000 & 0 & 0 & 1.000 \\

Qwen-2.5 7B & ARC & 1000 & 1000 & 0 & 0 & 1.000 \\
Qwen-2.5 7B & CSQA & 1000 & 999 & 0 & 1 & 0.999 \\
Qwen-2.5 7B & HellaSwag & 1000 & 1000 & 0 & 0 & 1.000 \\
Qwen-2.5 7B & MMLU & 1000 & 1000 & 0 & 0 & 1.000 \\
\bottomrule
\end{tabular}
\caption{Response format validity across models and datasets. Invalid responses are extremely rare, with valid ratios above 99.9\%.}
\label{tab:response_format_validity}
\end{table*}

\section{GPQA Baseline Extension}
\label{app:gpqa_baseline}

We additionally evaluate whether the pooled MPD analysis can be applied to
GPQA, a challenging expert-level MCQA benchmark. We use existing baseline
probability outputs for the complete set of 448 GPQA items from Gemma-2 9B,
Llama-3 8B, and Qwen-2.5 7B. Pooling the item representations across the three
models produces $N=1{,}344$ $(\text{item},\text{model})$ points. We then
compute MPD over all within-model and cross-model pairs using the same
aggregation defined in Section~\ref{sec:methodology}.

\begin{table}[h]
\centering
\small
\begin{tabular}{lrrr}
\toprule
Benchmark & Items & Pooled points & MPD \\
\midrule
GPQA & 448 & 1,344 & 0.477 \\
\bottomrule
\end{tabular}
\caption{Baseline pooled MPD for GPQA across three models. The value is
computed from existing model outputs rather than from additional inference.}
\label{tab:gpqa_baseline}
\end{table}

\section{Cross-Model \texorpdfstring{$H_{norm}$}{H\_norm} Consistency} \label{app:hnorm_consistency}
\begin{table}[h]
\centering
\small
\setlength{\tabcolsep}{4pt}
\resizebox{\columnwidth}{!}{
\begin{tabular}{lcc}
\toprule
Benchmark & Small-model mean $r$ & All-pair mean $r$ \\
\midrule
ARC       & 0.089 & $-0.010$ \\
CSQA      & 0.177 &  0.087 \\
HellaSwag & 0.102 &  0.056 \\
MMLU      & 0.028 &  0.066 \\
\bottomrule
\end{tabular}
}
\caption{Cross-model Spearman rank correlation of item-level $H_{norm}$.
\textit{Small-model mean $r$} averages over the three small-scale model pairs
(Gemma-2 9B, Llama-3 8B, Qwen-2.5 7B), while
\textit{All-pair mean $r$} averages over all 15 model pairs including large-scale
models. Near-zero values indicate that item-level $H_{norm}$ ranks are not
reproducible across models.}
\label{tab:hnorm_spearman}
\end{table}
To assess whether item-level $H_{norm}$ carries benchmark-intrinsic information
beyond model-specific calibration, we compute pairwise Spearman rank correlations
of item-level $H_{norm}$ between all model pairs across the four benchmarks.
Table~\ref{tab:hnorm_spearman} reports the resulting cross-model rank
correlations.

The small-model mean $r$ values range from 0.028 (MMLU) to 0.177 (CSQA), and the
all-pair mean $r$ drops to near zero or below for most benchmarks (ARC: $-$0.010).
This indicates that, while aggregate benchmark-level $H_{norm}$ rankings are
broadly stable across the well-calibrated models evaluated
(Table~\ref{tab:hnorm_aggregate}), the item-level ordering is
largely model-specific and cannot be used as a reliable per-item diagnostic.

\section{Taxonomy Annotation Agreement} \label{app:taxonomy_agreement}

\begin{table}[h]
\centering
\small
\resizebox{\columnwidth}{!}{
\begin{tabular}{lrrrr}
\toprule
Dataset & $n$ & $\kappa$ (A) & $\kappa$ (B) & Agreement (A/B) \\
\midrule
MMLU      & 14 & 1.000 & 1.000 & 100\% / 100\% \\
ARC       &  5 & undefined & undefined & 100\% / 100\% \\
CSQA      &  5 & 1.000 & 1.000 & 100\% / 100\% \\
HellaSwag &  5 & 0.250 & 0.545 & 40\% / 80\% \\
\bottomrule
\end{tabular}
}
\caption{Agreement between the author labels and two independent annotators.
For ARC, Cohen's $\kappa$ is undefined because all items received the X1 label,
leaving no category variance.}
\label{tab:taxonomy_iaa}
\end{table}

Agreement is high for MMLU, ARC, and CSQA, but substantially weaker for the
fine-grained HellaSwag subtype labels. For HellaSwag, the reported
author--annotator $\kappa$ values are 0.250 and 0.545 on a five-item subset.
Accordingly, we treat the fine-grained subtypes as qualitative categories and
base the primary reliability claim on the coarser Actionable-versus-Excluded
distinction. These results should be interpreted cautiously because the
double-annotated subsets are small.

\section{Noise Injection Taxonomy: Qualitative Examples} \label{app:taxonomy_examples}

This appendix presents representative examples for each failure type identified in Section~\ref{sec:taxonomy}. For excluded types (X1, X2) one example is shown to illustrate why they are excluded; for actionable failure types (A2--A5) one example is shown per type.

\subsection*{X1. Negation/EXCEPT Format}

\begin{PromptBox}{ARC \#997}
\footnotesize

\textbf{Question.}
\emph{``Which of these is not part of an atom?''}

\smallskip
\textbf{Choices.}

A. Paris\\
B. \textbf{isotope} {\scriptsize [GT]}\\
C. Seoul\\
D. Tokyo

\smallskip
{\scriptsize
\textcolor{gray}{
Original distractors: A = proton, C = nucleus, D = electron.
}
}

\end{PromptBox}
The question asks which option is not part of an atom. Under noise injection, the city names (Paris, Seoul, Tokyo) are also not parts of an atom, so the model cannot distinguish the intended answer (isotope) from the noise choices. Noise injection removes the semantically contrastive set that NOT/EXCEPT questions require, so failure here reflects a method limitation rather than an item flaw.

\subsection*{X2. Semantic Collision}

\begin{PromptBox}{CSQA \#659}
\footnotesize

\textbf{Question.}
\emph{``They wanted to try blowfish, so they went to get some where?''}

\smallskip
\textbf{Choices.}

A. Paris\\
B. \textbf{fish market} {\scriptsize [GT]}\\
C. Seoul\\
D. Tokyo {\scriptsize\textcolor{gray}{[semantic collision]}}

\end{PromptBox}
Tokyo is a semantically valid answer to this question because blowfish
(\textit{fugu}) is a well-known Japanese delicacy associated with Tokyo.
The injected city name happens to be correct, so model failures here reflect
the injection method rather than a problem with the item.
\subsection*{A3. Encoding Artifact}

\begin{PromptBox}{MMLU \#344}
\footnotesize

\textbf{Question.}
\emph{``Mr. Jones rolls a six-sided cube numbered 1, 2, 3, 4, 5, 6.
What is the probability he rolls a three?''}

\smallskip
\textbf{Choices.}

A. \textbf{6-Jan} {\scriptsize [GT]}\\
B. 5-Jan\\
C. 3-Jan\\
D. 2-Jan

\smallskip
{\scriptsize
\textcolor{gray}{
Intended fractions: A = $1/6$, B = $1/5$, C = $1/3$, D = $1/2$.
}
}

\end{PromptBox}
The answer choices are fractions (1/6, 1/5, 1/3, 1/2) that were automatically converted to date strings by spreadsheet software (e.g., $1/6 \to$ \textit{``6-Jan''}). After noise injection, these opaque strings are indistinguishable from nonsensical tokens; models consistently reject the ground truth choice ``6-Jan'' in favor of city-name distractors. This item is also flagged as \texttt{bad\_options\_clarity} by the LLM-based classifier.

\subsection*{A2. Answer-Key Error}

\begin{PromptBox}{MMLU \#428}
\footnotesize

\textbf{Question.}
\emph{``Which of the following is a biotic factor that could affect the
growth rate of a population?''}

\smallskip
\textbf{Choices.}

A. Volcanic eruption\\
B. Glacier melting\\
C. \textbf{Destruction of the ozone layer} {\scriptsize [GT in dataset]}\\
D. Sudden reduction in the animal food resource\\
\hspace*{1.6em}{\scriptsize\textcolor{gray}{[correct answer]}}

\end{PromptBox}
The ground truth labels option C (ozone layer destruction) as a biotic factor, but the ozone layer is an abiotic component of the environment. Option D (reduction in animal food resource) is the correct biotic factor. Under noise injection, the model correctly rejects the ground truth in favor of its domain knowledge, exposing the mislabel.

\subsection*{A5. Trivia-Dependent Underspecification}

\begin{PromptBox}{CSQA \#309}
\footnotesize

\textbf{Question.}
\emph{``Kramer wrote a self-referential book. What might that book be about?''}

\smallskip
\textbf{Choices.}

A. counter\\
B. \textbf{coffee table} {\scriptsize [GT]}\\
C. backpack\\
D. bedside table

\end{PromptBox}
The correct answer refers to a specific episode of \textit{Seinfeld} (Season 7, Episode 11) in which the character Kramer writes a coffee table book about coffee tables. Without this episode-specific knowledge, any furniture item is equally plausible, and after noise injection all remaining options (A, C, D) are equally uninformative. Models fail because no general semantic reasoning can resolve the question.

\subsection*{A4. Narrative Ambiguity}

\begin{PromptBox}{HellaSwag \#3001}
\footnotesize

\textbf{Context.}
\emph{``People are seen walking around a track and sitting down. One stands
before a track and looks off into the distance. he''}

\smallskip
\textbf{Candidate continuations.}

\noindent\hangindent=1.6em\hangafter=1
A. throws a ball up into a pit.\par

\noindent\hangindent=1.6em\hangafter=1
B. lifts up his arm and walks off the track.\par

\noindent\hangindent=1.6em\hangafter=1
C. yells to an audience and raises his arms up.\par

\noindent\hangindent=1.6em\hangafter=1
D. \textbf{runs down the track and into a sand pit.} {\scriptsize [GT]}\par

\end{PromptBox}
After replacing A, B, C with city names, only D (runs down the track and into a sand pit) remains as a real continuation option. The model still fails to select D, because the general track-and-field scene does not strongly cue the specific long-jump action, and the model assigns insufficient probability to D over the city-name options.

\section{LLM-Based Structural Classification} \label{app:llm_classification}

To provide an independent signal complementary to noise injection, we apply an MMLU-Redux-style structural classifier to all three benchmarks. The classifier uses GPT-4o-mini via the Openrouter\footnote{\url{https://openrouter.ai/}} (temperature 0) and assigns each item to one of three categories: \texttt{ok}, \texttt{bad\_question\_clarity} (structural defect in the question stem), or \texttt{bad\_options\_clarity} (formatting defect in the answer choices). Wrong-answer and multiple-correct-answer judgments are excluded because they require domain knowledge and are prone to model-specific false positives.
Table~\ref{tab:llm_classification} reports the resulting structural classification outcomes across datasets.

\begin{table}[H]
\centering
\small
\begin{tabular}{lrrrrr}
\toprule
Dataset & Total & ok & \texttt{bq} & \texttt{bo} & Flag rate \\
\midrule
ARC             & 997  & 989 &  6 &  2 & 0.8\% \\
CommonsenseQA   & 1000 & 816 & 95 & 89 & 18.4\% \\
MMLU            & 1000 & 972 & 15 & 13 & 2.8\% \\
\bottomrule
\end{tabular}
\caption{LLM-based structural classification results (v6). \texttt{bq} = \texttt{bad\_question\_clarity}; \texttt{bo} = \texttt{bad\_options\_clarity}.}
\label{tab:llm_classification}
\end{table}

The high flag rate for CommonsenseQA (18.4\%) reflects crowdsourcing artifacts: pronoun ambiguity errors (\textit{``Where did it put it?''} with \textit{it} as subject), ConceptNet-generated distractors that appear as meaningless tokens (\textit{``wisconsin''}, \textit{``hatred''}), and incomplete sentence stems. MMLU flags (2.8\%) concentrate in PDF/CSV extraction artifacts: fractions converted to Excel date strings (e.g., $1/6 \to$ \textit{``6-Jan''}), superscripts replaced with \textit{``?''}, and duplicated answer options. The MMLU encoding artifact pattern directly corresponds to the A3 encoding-artifact type identified in the noise-injection taxonomy, providing convergent evidence from two independent methods.

\subsection*{Classification Procedure}

Classification follows a strict five-step decision hierarchy applied in order: (1) question clarity, (2) options clarity, (3--5) correctness checks (not used for flagging). Items are flagged only if a structural defect is unambiguous without domain knowledge. A post-processing consistency rule corrects 77 self-contradictory outputs where the step reasoning concluded ``ok'' but the \texttt{error\_type} field returned a flag; in such cases the \texttt{error\_type} is overridden to \texttt{ok}.

\end{document}